\documentclass[letterpaper]{article} 
\usepackage[]{aaai2026}  
\usepackage{times}  
\usepackage{helvet}  
\usepackage{courier}  
\usepackage[hyphens]{url}  
\usepackage{graphicx} 
\usepackage{natbib}  
\usepackage{caption} 
\usepackage{algorithm}
\usepackage{algorithmic}

\usepackage{booktabs}     
\usepackage{multirow}     
\usepackage{amsmath}      
\usepackage{newfloat}
\usepackage{listings}
\DeclareCaptionStyle{ruled}{labelfont=normalfont,labelsep=colon,strut=off} 
\floatstyle{ruled}
\newfloat{listing}{tb}{lst}{}
\floatname{listing}{Listing}
\title{Coupled Degradation Modeling and Fusion: A VLM-Guided Degradation-Coupled Network for Degradation-Aware Infrared and Visible Image Fusion}
\author{
     Tianpei Zhang, Jufeng Zhao, Yiming Zhu, Guangmang Cui
}
\usepackage{bibentry}

\begin{document}

\maketitle
\begin{abstract}
Existing Infrared and Visible Image Fusion (IVIF) methods typically assume high-quality inputs. However, when handing degraded images, these methods heavily rely on manually switching between different pre-processing techniques. This decoupling of degradation handling and image fusion leads to significant performance degradation. In this paper, we propose a novel \textbf{V}LM-\textbf{G}uided \textbf{D}egradation-\textbf{C}oupled Fusion network (VGDCFusion), which tightly couples degradation modeling with the fusion process and leverages vision-language models (VLMs) for degradation-aware perception and guided suppression. Specifically, the proposed Specific-Prompt Degradation-Coupled Extractor (SPDCE) enables modality-specific degradation awareness and establishes a joint modeling of degradation suppression and intra-modal feature extraction. In parallel, the Joint-Prompt Degradation-Coupled Fusion (JPDCF) facilitates cross-modal degradation perception and couples residual degradation filtering with complementary cross-modal feature fusion. Extensive experiments demonstrate that our VGDCFusion significantly outperforms existing state-of-the-art fusion approaches under various degraded image scenarios. Our code is available at https://github.com/Lmmh058/VGDCFusion.
\end{abstract}


\section{Introduction}
Infrared and visible images offer complementary information due to their distinct imaging mechanisms: infrared captures thermal radiation but lacks texture, while visible images provide rich details and color but are sensitive to lighting and occlusion. Infrared and Visible Image Fusion(IVIF) integrates these modalities to produce images that combine the strengths of both, enhancing the performance of downstream tasks such as autonomous driving \cite{abrecht2024deep}, object detection \cite{hu2025datransnet}.

\begin{figure}[!t]
\centering
\includegraphics[width=0.85\columnwidth]{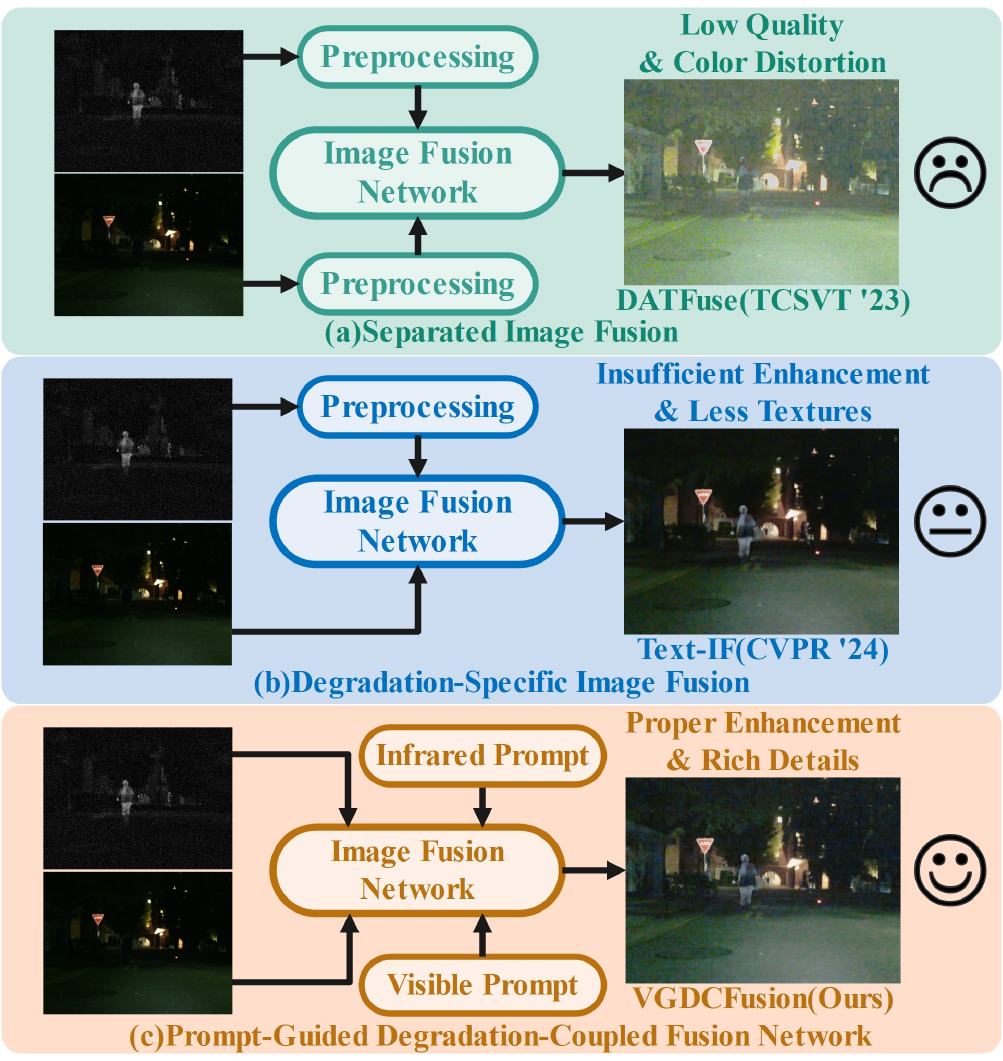}
\caption{
Comparison of fusion strategies under dual-modality degradation.}
\label{Fig1}
\end{figure}

In real-world applications, input images may face degradation. For instance, visible images may suffer from low light or overexposure, while infrared images frequently face low contrast or noise. Such degradations severely compromise the performance of IVIF methods: a blurry low-light visible image and a noisy infrared image, when fused naively, can result in lost target details or misleading artifacts, directly hindering downstream tasks. Thus, developing IVIF methods that are inherently robust to diverse degradations is critical for practical application.

\begin{figure*}[!t]
\centering
\includegraphics[width=0.85\textwidth]{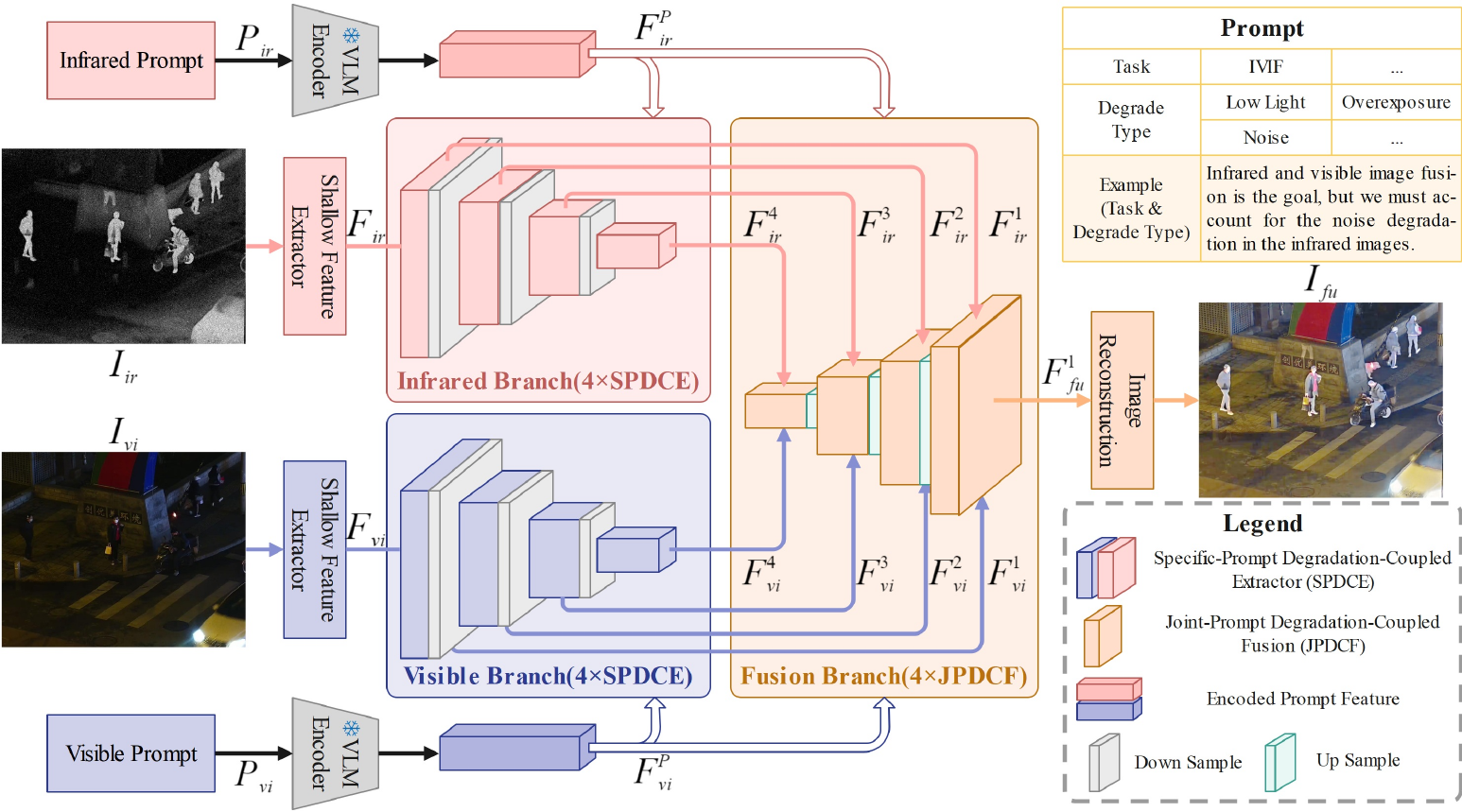}
\caption{
Architecture of VGDCFusion. The top‑right inset shows the prompt components (where "IVIF" denotes Infrared and Visible Image Fusion) and an example; the bottom‑right legend details each block in the diagram. Notably, the encoded prompt features are injected into every layer of both SPDCE and JPDCF modules to guide degradation‑aware fusion.
} 
\label{Network}
\end{figure*}

Most recent IVIF methods \cite{xu2022cufd,tang2023datfuse,xie2024fusionmamba,zhang2025exploring} assume high-quality inputs and tend to retain noticeable degradation artifacts when applied to degraded images. To address the issue of degraded fusion results caused by input artifacts, existing approaches often rely on complex preprocessing pipelines to restore source images before fusion. Meanwhile, recent studies have tackled specific degradation types or single-modality degradation. For instance, LVIF-Net \cite{chang2024lvif} targets low-light images, while Text-IF \cite{yi2024textif} employs vision-language models (VLMs) for text-guided fusion. However, these strategies remain limited in handling diverse and cross-modal degradations comprehensively. Specifically:
(1) Fusion quality is contingent upon the preprocessing enhancements and lacks synergy between degradation optimization and fusion, leading to performance degradation.
(2) Frequent switching of preprocessing methods for different degradations compromises the simplicity and practicality of fusion workflows.

Fig.\ref{Fig1} compares fusion strategies under dual-modality degradation: (a) Separated fusion applies modality-specific preprocessing to remove degradation; it yields low quality and distortion in the output; (b) Degradation-specific fusion networks handle a single degradation type; they may insufficiently enhance features and miss fine texture. To enhance both performance and usability in degraded scenarios, we identify two key challenges: 
(1) \textbf{Decoupling of degradation and fusion}: Current pipelines separate degradation handling from fusion, leading to misalignment between artifact suppression and feature preservation. A unified framework is needed to jointly optimize both.
(2) \textbf{Limited adaptability to cross-modal degradations}: Existing methods rely on manual adjustment for diverse degradations, lacking the ability to autonomously perceive and adapt to varied patterns.

Accordingly, we propose the VLM‑Guided Degradation-Coupled Fusion Network (VGDCFusion), with two modules designed for targeted solutions: Specific‑Prompt Degradation-Coupled Extractor (SPDCE) and Joint‑Prompt Degradation-Coupled Fusion  (JPDCF). SPDCE uses modality‑specific prompts for intra‑modal degradation awareness and multi‑scale local‑global extraction to mine features and suppress degradation. JPDCF leverages joint prompts to guide degradation perception and integrates residual degradation filtering with complementary feature aggregation across scales. As shown in Fig.\ref{Fig1}(c), our network’s coupling of degradation modeling and fusion process, combined with prompt guidance, yields high‑quality outputs with effective degradation removal. Our main contributions are summarized as follows:

\begin{itemize}
    \item To address the decoupling of degradation handling and fusion, we propose VGDCFusion, a novel framework that deeply couples degradation modeling with fusion via VLM prompt guidance, establishing a degradation-aware architecture.
    \item 
    We propose SPDCE, enabling intra‑modal degradation awareness and synergistically performing degradation suppression and feature extraction.
    \item 
    To handle cross-modal degradation adaptively, we propose JPDCF, which integrates cross-modal prompt guidance with residual artifact filtering and complementary feature fusion.
\end{itemize}

\section{Related Work}
\subsection{Infrared and Visible Image Fusion}
Deep learning has driven rapid prograss in IVIF, with methods based on autoencoder (AE) \cite{li2021rfn,xu2022cufd}, generative adversarial network (GAN) \cite{ma2019fusiongan,ma2020ganmcc}, convolutional neural network (CNN) \cite{liu2023sgfusion, xu2020u2fusion}, and Transformer \cite{tang2023datfuse,ma2022swinfusion}. However, these methods inherently assume high-quality inputs: they perform well on clean images but retain degradation artifacts (e.g., noise, blur) and lose critical details when applied to degraded inputs. 

Recent efforts to address degradation remain limited: LVIF‑Net \cite{chang2024lvif} for low‑light visible images, while Text‑IF \cite{yi2024textif} uses VLMs for text-guided fusion but handles only single-modality or single-type degradation. Neither can comprehensively tackle cross-modal, diverse degradations (e.g., low-light visible with noisy infrared)—a critical gap our work addresses.

\subsection{Vision Language Model}
Vision-language models (VLMs) jointly model visual and textual modalities and have been widely adopted in multi-modal tasks \cite{antol2015vqa}. Recent models like CLIP \cite{radford2021clip} and BLIP \cite{li2022blip}, empowered by large-scale pretraining, demonstrate strong visual–textual alignment and high-level semantic understanding. More recently, VLMs have been introduced into image enhancement and fusion, offering global context awareness and semantic guidance for feature extraction. For instance, Cai \textit{et al.} \cite{cai2025degradation} proposed a VLM-based degradation-aware enhancement network, while Text-IF \cite{yi2024textif} utilizes VLMs for interactive single-degradation image fusion.

\begin{figure}[!t]
\centering
\includegraphics[width=0.9\columnwidth]{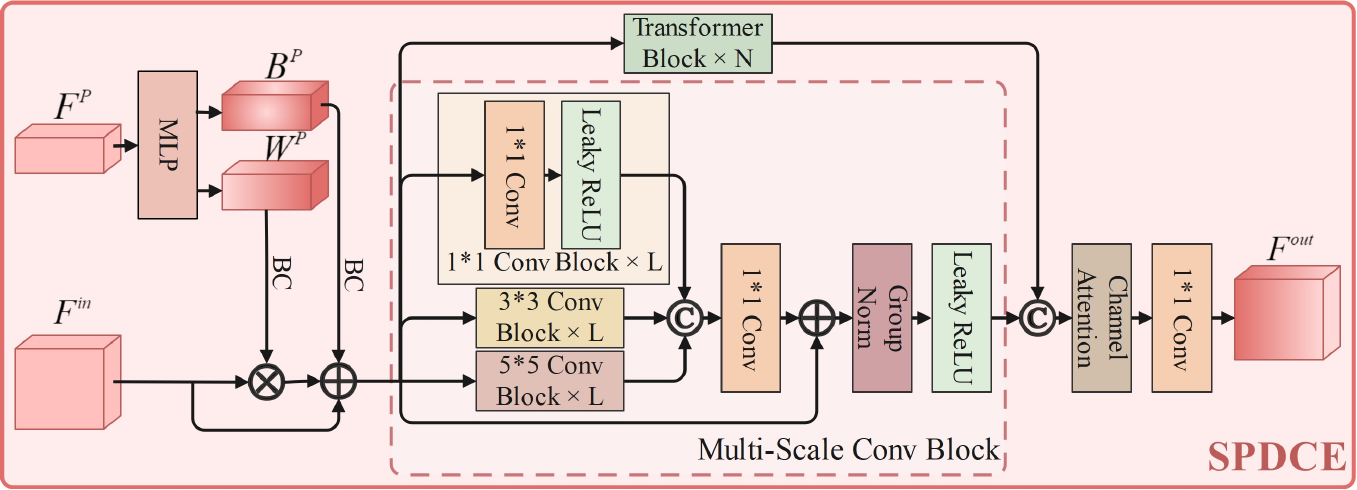}
\caption{Network architecture of the SPDCE. "BC" denotes the abbreviation for the broadcasting operation.}
\label{Network_SPDCE}
\end{figure}

\section{Method}
\subsection{Overall Framework}
The architecture of VGDCFusion is illustrated in Fig.~\ref{Network}. The degraded infrared image $I_{ir}$ is projected into feature space via a 3×3 convolution to produce shallow features $F_{ir}=SF(I_{ir})$. Concurrently, the infrared prompt $P_{ir}$ is encoded by a frozen VLM encoder into prompt features $F^P_{ir}=VLM(P_{ir})$. The infrared features $F_{ir}$ and the corresponding prompt features $F^{P}_{ir}$ are then passed through four successive SPDCE modules for intra-modal degradation awareness, suppression, and multi-scale feature mining.
{\footnotesize
\begin{equation}
\label{eq:SPDCE}
\begin{aligned}
F^{1}_{ir} &= SPDCE(F_{ir},F^{P}_{ir})\\
F^{n}_{ir} &= SPDCE(DS(F^{n-1}_{ir}),F^{P}_{ir}),n\in\{2,3,4\}  
\end{aligned}
\end{equation}}
where $DS(\cdot)$ denotes downsampling, and $F^{n}_{ir}$ are the infrared features after $n$ SPDCE layers. The visible features $F^{n}_{vi}$ are obtained similarly via SPDCE for feature extraction and degradation suppression. The image features $(F^{n}_{ir},F^{n}_{vi})$ and prompt features $(F^{P}_{ir},F^{P}_{vi})$ are then passed through four sequential JPDCF modules for complementary feature aggregation and residual degradation filtering guided by cross‑modal degradation awareness.
{\footnotesize
\begin{equation}
\label{eq:JPDCF}
\begin{aligned}
F^{m}_{ir} &= JPDCF(US(F^{m+1}_{fu}),F^{m}_{ir},F^{m}_{vi},F^{P}_{ir},F^{P}_{vi})\\
F^{4}_{fu} &= JPDCF(F^{4}_{ir},F^{4}_{vi},F^{P}_{ir},F^{P}_{vi}),m\in\{1,2,3\}  
\end{aligned}
\end{equation}}
where $US(\cdot)$ denotes upsampling, and $F^{m}_{fu}$ represents the fused features after $5 - m$ JPDCF modules. The feature $F^{1}_{fu}$ is fed into a reconstruction module $Re(\cdot)$ comprising three successive 3×3 convolutions each followed by Leaky ReLU, yielding the fused image $I_{fu} = Re(F^{1}_{fu})$.

\subsection{Specific-Prompt Degradation-Coupled Extractor}
To achieve coupling between intra-modal degradation modeling and feature extraction, as well as modality-specific degradation awareness, we design the SPDCE, as illustrated in Fig.~\ref{Network_SPDCE}. Prompt features $F^P$ guide the input image features to embed degradation-related cues, enhancing the module’s sensitivity to degradation, which can be formulated as:
{\footnotesize
\begin{equation}
\label{eq:SPDCE_Guidance}
\begin{aligned}
 W^{P},B^{P} &= MLP(F^{P})\\
 F^{G} &= F^{in} \times BC(W^{P}) + BC(B^{P}) + F^{in}
\end{aligned}
\end{equation}}
where $W^{P}$ and $B^{P}$ denote the weights and bias used for feature guidance, $MLP(\cdot)$ is a multilayer perceptron, and $BC(\cdot)$ represents feature broadcasting. 

The guided image features $F^{G}$ are then processed by a multi-scale convolution block $MSConv(\cdot)$ to multi-scale local feature extraction, and a series of $N=2$ Transformer blocks $Trm(\cdot)$ for long-range dependency modeling, while jointly suppressing degradation across scales. Finally, a channel attention module $CA(\cdot)$ and a 1×1 convolution $Conv_{1}(\cdot)$ are applied to emphasize informative channels and aggregate channel-wise information, producing the SPDCE output feature $F^{out}$. This process is defined as:
{\footnotesize
\begin{equation}
\label{eq:SPDCE_Extraction}
\begin{aligned}
 F' &= Cat(Trm(F^{G}),MSConv(F^{G}))\\
 F^{out} &= Conv_{1}(CA(F'))
\end{aligned}
\end{equation}}
where \({Cat}(\cdot)\) denotes channel-wise concatenation.

\begin{figure}[!t]
\centering
\includegraphics[width=0.8\columnwidth]{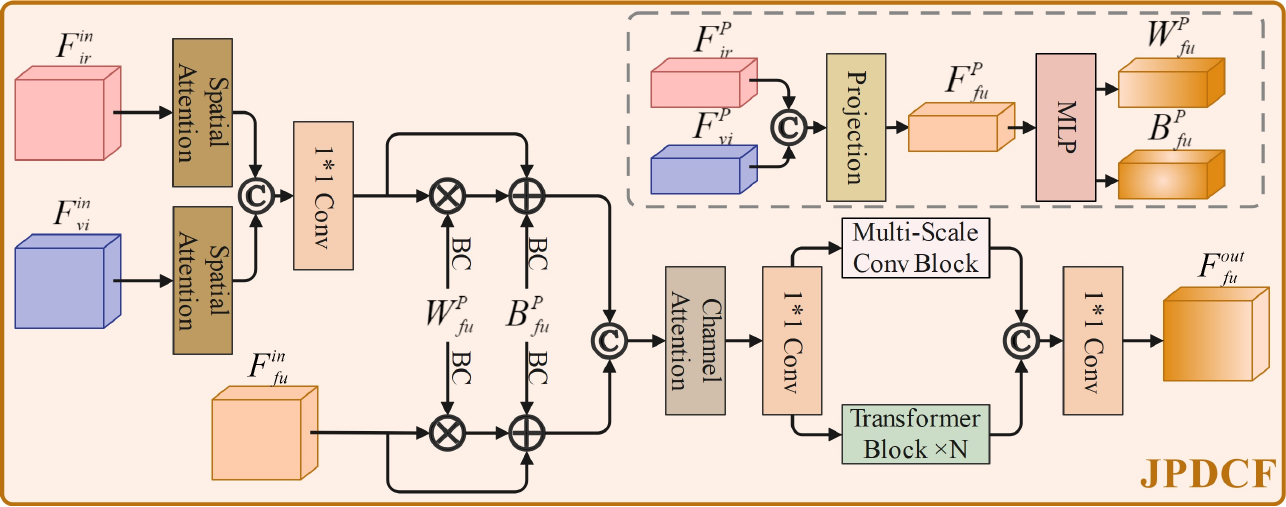}
\caption{Network architecture of the JPDCF. The computation of $W^{P}_{fu}$ and $B^{P}_{fu}$ is shown in the grey box (top right).}
\label{Network_JPDCF}
\end{figure}

\subsubsection{Multi-Scale Conv Block:}
To complement the global features extracted by the Transformer with local information, we design a Multi-Scale Conv Block that captures local features and suppresses degradation under varying receptive fields. Specifically, the input features $F^G$ are passed through three parallel convolutional branches with kernel sizes of 1, 3, and 5. Each branch consists of $L=3$ convolution layers followed by LeakyReLU activation $LReLU(\cdot)$. A subsequent 1×1 convolution aggregates channel-wise information, while a residual connection preserves the original features. Finally, group normalization $GN(\cdot)$ and activation are applied to produce the output local features $F^{local}$.
{\footnotesize
\begin{equation}
\label{eq:Multi-Scale Conv Block}
\begin{aligned}
 F_{k} &= ConvBlock^{L}_{k}(F^{G}), k \in \{1,3,5\}\\
 F' &= Cat(F_{1},F_{3},F_{5})\\
 F^{local} &= LReLU(GN(Conv_{1}(F')+F^{G}))
\end{aligned}
\end{equation}}
where $ConvBlock^{L}_{k}(\cdot)$ represents a sequence of $L$ convolutional blocks with kernel size $k$.

\subsection{Joint-Prompt Degradation-Coupled Fusion}
To establish an effective coupling between degradation filtering and cross-modal feature fusion, and to enable degradation awareness during fusion, we design the JPDCF module. We first fuse prompt features from both modalities $F^{P}_{ir},F^{P}_{vi}$ to enable more comprehensive prompt guidance. This process is defined as:
{\footnotesize
\begin{equation}
\label{eq:Prompt Fusion}
\begin{aligned}
 F^{P}_{fu} &= Proj(Cat(F^{P}_{ir},F^{P}_{vi}))\\
 W^{P}_{fu}, B^{P}_{fu} &= MLP(F^{P}_{fu})\\
\end{aligned}
\end{equation}}
where $Proj(\cdot)$ denotes the feature projection operation, $F^{P}_{fu}$ represents the fused prompt features, and $W^{P}_{fu}, B^{P}_{fu}$ are the weights and biases used in prompt guidance. The modality features $F^{in}_{ir}$ and $F^{in}_{vi}$ from SPDCE are first processed by a spatial attention module $SA(\cdot)$ to emphasize informative spatial regions. These features are then concatenated along the channel dimension and fused via a 1×1 convolution. The resulting features are combined with the prompt-guided degradation-aware embedding computed using $W^{P}_{fu}, B^{P}_{fu}$, and further concatenated with the guided output from the previous JPDCF layer. Channel attention and aggregation are then applied, followed by a multi-scale convolution block and a cascade of $N$ Transformer blocks to filter residual degradation and enhance cross-modal feature fusion. Finally, the features are fused via channel-wise concatenation and a 1×1 convolution to produce the JPDCF output $F^{out}_{fu}$. The overall process is formulated as:
{\footnotesize
\begin{equation}
\label{eq:JPDCF_Fusion}
\begin{aligned}
 F' &= Conv_{1}(Cat(SA(F^{in}_{ir}),SA(F^{in}_{vi})))\\
 F'' &= F' \times BC(W_{fu}^{P}) + BC(B_{fu}^{P}) + F'\\ 
 F''' &= F_{fu}^{in} \times BC(W_{fu}^{P}) + BC(B_{fu}^{P}) + F_{fu}^{in}\\
 F_{fu} &= Conv_{1}(CA(Cat(F'',F''')))\\
 F^{out}_{fu} &= Conv_{1}(Cat(MSConv(F_{fu}),Trm(F_{fu})))
\end{aligned}
\end{equation}}
\subsection{Loss Function Design and Analysis}
To ensure that VGDCFusion preserves sufficient intensity and rich texture while avoiding color distortion in the fused results, we design the overall loss function as a combination of three components: intensity loss $L_{int}$, texture loss $L_{texture}$, and color loss $L_{color}$.
{\footnotesize
\begin{equation}
\label{eq:Loss Function}
 L = \alpha L_{int} + \beta L_{texture} + \gamma L_{color}
\end{equation}}
where $\alpha$, $\beta$, and $\gamma$ are the weighting hyperparameters for the three loss terms. Specifically, the intensity loss encourages the fused image to retain salient structures by computing the difference between the fused image and the maximum pixel intensity of the source images. This is formulated as:
{\footnotesize
\begin{equation}
\label{eq:Int loss}
 L_{int} = \frac{1}{HW}||I_{fu}-max(I^{ref}_{ir},I^{ref}_{vi})||_{1}
\end{equation}}
where, $H$ and $W$ denote the height and width of the image, respectively; $I_{fu}$, $I^{ref}_{ir}$, and $I^{ref}_{vi}$ represent the fused image and the non-degraded infrared and visible reference images. The operator $max(\cdot)$ selects the element-wise maximum. Next, the texture loss is computed as the difference between the fused image and the maximum gradient magnitude of the source images, encouraging the preservation of fine texture details. It is defined as:
{\footnotesize
\begin{equation}
\label{eq:Texture loss}
 L_{texture} = \frac{1}{HW}||\nabla I_{fu}-max(\nabla I^{ref}_{ir},\nabla I^{ref}_{vi})||_{1}
\end{equation}}
where, $\nabla$ denotes the Sobel gradient operator.
Finally, the color loss is computed as the difference between the fused image and the non-degraded visible reference image in the Cb and Cr channels of the YCbCr color space. This loss mitigates potential color distortion introduced during degraded fusion and is defined as:
{\footnotesize
\begin{equation}
\label{eq:Color loss}
\begin{aligned}
L_{color} = \frac{1}{HW} \sum_{C \in \{Cb, Cr\}} \left\| C_{fu} - C^{ref}_{vi} \right\|_1
\end{aligned}
\end{equation}}
where $C$ denotes the Cb or Cr channel image.

\begin{table}[!t]  
\centering
{\small
\begin{tabular}{ll}
\toprule[1pt]
Methods(\textit{Source \& Year \& Type})     \\ \hline
GANMcC \cite{ma2020ganmcc} \textit{TIM'2020, GAN-based} \\ 
SDNet \cite{zhang2021sdnet} \textit{IJCV'2021, CNN-based} \\ 
CSF \cite{xu2021classification} \textit{TCI'2021, AE-based} \\ 
RFN-Nest \cite{li2021rfn} \textit{INFFUS'2021, AE-based} \\ 
DATFuse \cite{tang2023datfuse} \textit{TCSVT'2023, Transformer-based}  \\ 
ITFuse \cite{tang2024itfuse} \textit{PR'2024, Transformer-based}  \\
Text-IF \cite{yi2024textif} \textit{CVPR'2024, Transformer-based}  \\
\hline
$\star$ \textbf{VGDCFusion(Ours), Transformer-based} \\
\bottomrule[1pt]
\end{tabular}}
\caption{Configurations for all comparative experiments}
\label{tab:comparative methods}
\end{table}

\begin{figure*}[!t]
\centering
\includegraphics[width=0.9\textwidth]{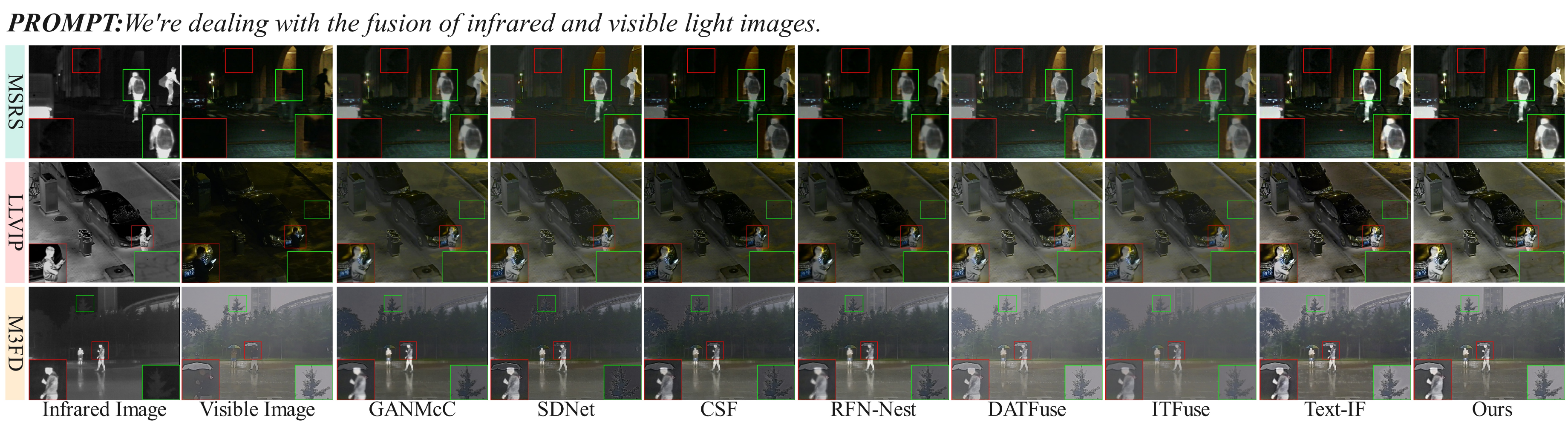}
\caption{
Qualitative comparison of VGDCFusion and seven comparative methods on MSRS, LLVIP, and M3FD datasets. “Prompt” denotes the fusion guidance in VGDCFusion. Key regions are highlighted with red/green boxes for visual clarity.}
\label{Qualitative_Normal}
\end{figure*}

\begin{table*}[!t]
\centering
{\small
\begin{tabular}{c|cccccc}
\toprule
\multirow{2}{*}{\textbf{Method}} 
& \multicolumn{6}{c}{\textbf{Infrared and Visible Image Fusion without degraded guidance(MSRS/LLVIP/M3FD)}} \\
\cline{2-7}
& \textbf{FMI$\uparrow$} & \textbf{MI$\uparrow$} & \textbf{Q$_{abf}\uparrow$} & \textbf{Q$_{p}\uparrow$} & \textbf{Q$_{C}\uparrow$} & \textbf{Q$_{W}\uparrow$}\\
\midrule
GANMcC     
& 0.937/0.906/0.894 & 1.461/1.688/1.699 & 0.344/0.260/0.341 & 0.248/0.271/0.296 & 0.599/0.593/0.623 & 0.593/0.483/0.589 \\
SDNet   
& 0.935/0.909/0.894 & 0.939/1.796/1.939  & 0.351/0.533/0.524 & 0.207/0.284/0.313 & 0.486/0.679/0.683 & 0.700/0.777/0.818 \\
CSF
& 0.940/0.907/0.894 & 1.196/1.666/1.803 & 0.299/0.420/0.469 & 0.233/0.335/0.350 & 0.452/0.565/0.657 & 0.619/0.626/0.729 \\
RFN-Nest
& 0.941/0.907/0.899 & 1.323/1.707/1.791 & 0.312/0.312/0.393 & 0.257/0.267/0.285 & 0.589/0.656/0.660 & 0.551/0.489/0.614 \\
DATFuse
& 0.943/0.907/0.884 & 1.927/2.191/2.425 & 0.561/0.470/0.494 & 0.359/0.295/0.365 & 0.619/0.622/0.701 & 0.847/0.684/0.588 \\
ITFuse
& 0.936/0.902/0.887 & 1.366/1.768/1.706 & 0.275/0.212/0.204 & 0.236/0.239/0.271 & 0.529/0.634/0.556 & 0.506/0.386/0.377 \\
Text-IF
& 0.951/0.919/0.912 & 2.059/1.929/2.068 & 0.609/0.662/0.660 & \textbf{0.410}/0.407/0.484 & 0.707/0.677/0.711 & 0.902/0.888/0.895 \\
\textbf{Ours}     
& \textbf{0.953}/\textbf{0.928}/\textbf{0.914} & \textbf{2.260}/\textbf{2.581}/\textbf{2.898} & \textbf{0.643}/\textbf{0.742}/\textbf{0.690} & 0.405/\textbf{0.462}/\textbf{0.491} & \textbf{0.752}/\textbf{0.841}/\textbf{0.739} & \textbf{0.916}/\textbf{0.921}/\textbf{0.898} \\
\bottomrule
\end{tabular}
}
\caption{
Quantitative comparison of VGDCFusion and seven comparative methods on MSRS, LLVIP, and M3FD datasets. $\uparrow$ indicates that higher values denote better performance. The best-performing results for each metric are highlighted in bold.}
\label{tab:Quantitative_Normal}
\end{table*}

\section{Experiment}
\subsection{Experimental Settings}
\subsubsection{Implementation details:}
We implemented two VGDCFusion variants: a degradation-aware version (with prompts) and a degradation-agnostic version (without prompts). For the former, 2,278 images were randomly selected from the EMS dataset \cite{yi2024textif}, which includes low-light/overexposed visible and low-contrast/noisy infrared images. For the latter, 11,025 images were sampled from the LLVIP dataset \cite{jia2021llvip}. All training images were cropped into 96×96 patches and fed to the network with a batch size of 16. The models were trained for 150 and 100 epochs, respectively, using a learning rate of 2.5e-4. Loss weights $\alpha$, $\beta$, and $\gamma$ were set to 10, 12, and 10. Experiments were conducted on an NVIDIA GeForce RTX 4090D.

\subsubsection{Datasets:}
To evaluate the degradation-aware model under challenging conditions, we randomly selected 69, 69, 35, and 35 image pairs from the EMS dataset \cite{yi2024textif}, corresponding to four degradation types: low-light \& low-contrast, low-light \& noise, overexposure \& low-contrast, and overexposure \& noise. For the degradation-agnostic model, 45, 50, and 49 images were sampled from the MSRS \cite{tang2022piafusion}, LLVIP \cite{jia2021llvip}, and M3FD \cite{liu2022target} datasets, respectively, to assess the network’s generalization and superiority.

\subsubsection{Comparative Methods and Metrics:}
Details of the comparative fusion methods are summarized in Table~\ref{tab:comparative methods}. For the degradation-aware model, due to the absence of "perfect reference images", we adopt four widely used no-reference metrics—AG \cite{cui2015detail}, EI \cite{rajalingam2018hybrid}, SD \cite{rao1997SD}, and SF \cite{eskicioglu1995image}—for fair and quantitative evaluation of fusion quality under degradation. For the degradation-agnostic model, we use six commonly adopted metrics—FMI \cite{haghighat2011non,qu2002information}, MI  \cite{qu2002information}, $Q_{abf}$ \cite{xydeas2000objective}, $Q_P$ \cite{zhao2007performance}, $Q_C$ \cite{cvejic2005qc}, and $Q_W$ \cite{piella2003new}—to quantitatively compare fusion performance. For all ten metrics, higher values indicate better performance.

\subsection{Image Fusion without Degrade Guidance}
\subsubsection{Qualitative Analysis:}
Fig.~\ref{Qualitative_Normal} shows the qualitative comparison between our degradation-agnostic model and seven competing methods. These comparative methods exhibit various limitations: (1) \textbf{Inadequate inheritance of infrared information:} GANMcC, CSF, RFN-Nest, and ITFuse fail to preserve salient infrared targets (highlighted by red boxes in rows 2–3) and detailed edge structures (green box in row 2). DATFuse struggles with retaining prominent targets (green box in row 1), while Text-IF fails to preserve infrared edge detail (red box in row 1). (2) \textbf{Poor retention of visible background:} Except for DATFuse and Text-IF, most methods over-rely on infrared intensity for background construction, leading to visually suboptimal results. (3) \textbf{Noticeable edge artifacts:} When fusing visible detail, SDNet, CSF, RFN-Nest, and Text-IF produce unnatural edge distortions.

\subsubsection{Quantitative Analysis:}
Tab.~\ref{tab:Quantitative_Normal} presents the quantitative comparison between our degradation-agnostic model and seven comparative approaches. Except for $Q_{P}$ on the MSRS dataset—where our method ranks second to Text-IF—our model consistently outperforms all competing methods across all other metrics on the MSRS, LLVIP, and M3FD datasets, demonstrating superior fusion performance even without explicit degradation guidance.

\begin{figure*}[!t]
\centering
\includegraphics[width=0.9\textwidth]{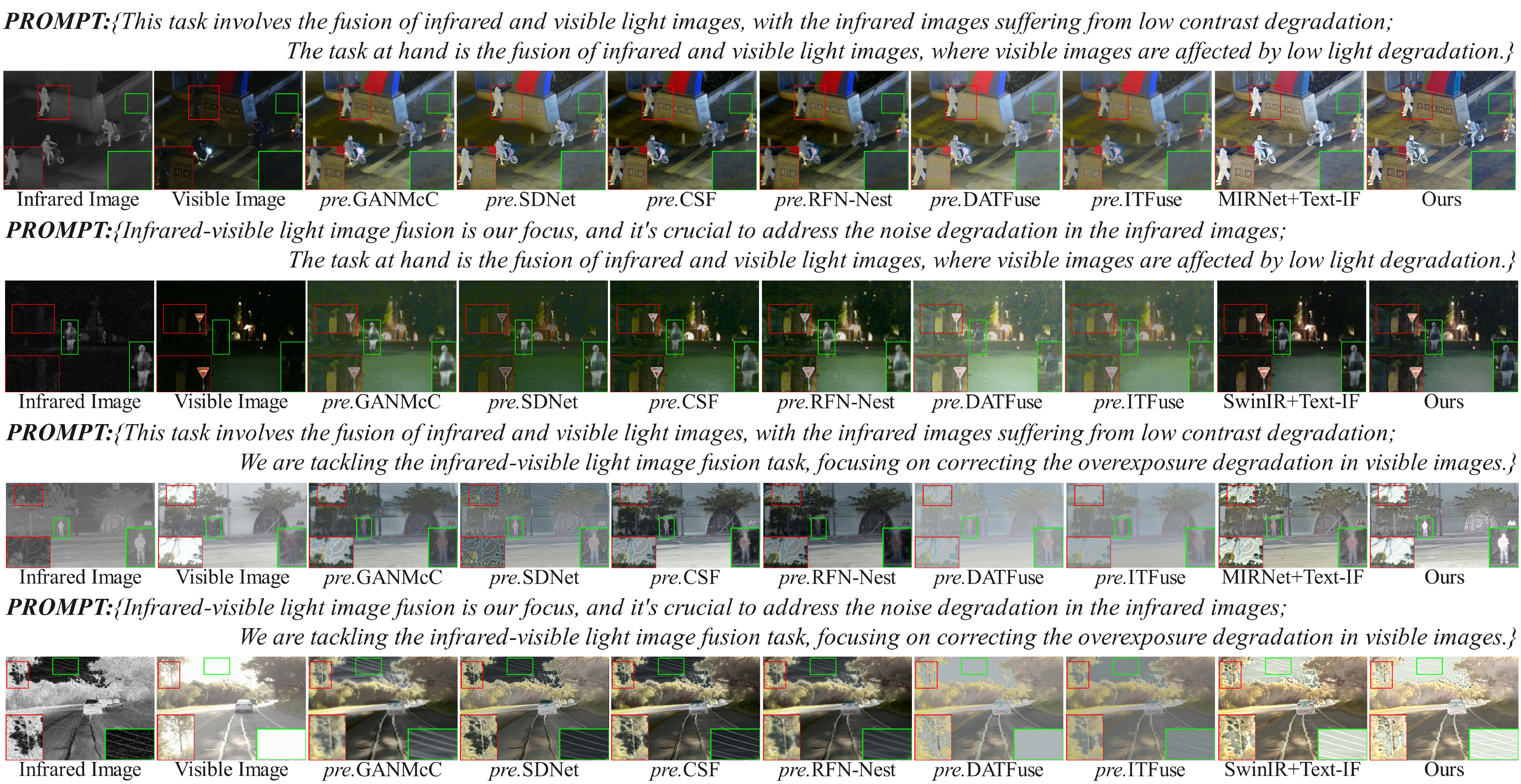}
\caption{
Qualitative comparison of degraded image fusion on the EMS dataset using VGDCFusion and seven comparative methods. Visible images suffer from low-light/overexposure, and infrared images from low-contrast/noise. $\textit{pre.}$ indicates preprocessing methods for degradation (IAT for visible brightness correction, MIRNet for IR contrast enhancement, and SwinIR for IR denoising). The “Prompt” sentences correspond to the infrared and visible degradation prompts used in VGDCFusion.}
\label{Qualitative_Degrade}
\end{figure*}

\begin{table*}[!t]
\centering
{\small
\begin{tabular}{c|cccc|cccc|cccc|cccc}
\toprule
\multirow{3}{*}{\textbf{Method}} 
& \multicolumn{16}{c}{\textbf{Infrared and Visible Image Fusion with degraded    guidance}} \\
\cline{2-17}
&\multicolumn{4}{c|}{\textbf{Low-light\&LowContrast}} & \multicolumn{4}{c|}{\textbf{Low-light\&Noise}} & \multicolumn{4}{c|}{\textbf{Overexposure\&LowContrast}} & \multicolumn{4}{c}{\textbf{Overexposure\&Noise}}\\

& \textbf{AG$\uparrow$} & \textbf{EI$\uparrow$} & \textbf{SD$\uparrow$} & \textbf{SF$\uparrow$} & \textbf{AG$\uparrow$} & \textbf{EI$\uparrow$} & \textbf{SD$\uparrow$} & \textbf{SF$\uparrow$} & \textbf{AG$\uparrow$} & \textbf{EI$\uparrow$} & \textbf{SD$\uparrow$} & \textbf{SF$\uparrow$} & \textbf{AG$\uparrow$} & \textbf{EI$\uparrow$} & \textbf{SD$\uparrow$} & \textbf{SF$\uparrow$}\\
\midrule
\textit{pre.}GANMcC    
& 3.036 & 32.130 & 33.993 & 8.450 & 2.800 & 29.653 & 32.522 & 8.102 & 3.938 & 41.532 & 43.071 & 10.308 & 3.676 & 39.259 & 47.325 & 9.299 \\
\textit{pre.}SDNet
& 4.294 & 44.907 & 30.182 & 12.768 & 3.589 & 37.598 & 27.921 & 11.573 & 5.551 & 57.502 & 30.225 & 15.017 & 5.556 & 58.234 & 46.848 & 14.965 \\
\textit{pre.}CSF
& 3.691 & 39.045 & 38.916 & 10.555 & 3.124 & 33.210 & 36.132 & 9.564 & 5.514 & 58.221 & 51.632 & 14.290 & 4.848 & 51.420 & 51.421 & 12.723 \\
\textit{pre.}RFN-Nest
& 3.051 & 32.860 & 38.560 & 8.341 & 2.789 & 30.029 & 36.855 & 7.802 & 4.667 & 50.330 & 51.265 & 12.272 & 3.758 & 40.712 & 49.532 & 9.372 \\
\textit{pre.}DATFuse
& 3.976 & 41.552 & 39.913 & 11.712 & 4.271 & 44.748 & \textbf{42.146} & 12.884 & 2.469 & 24.867 & 20.063 & 6.931 & 3.958 & 40.315 & 34.089 & 11.868 \\
\textit{pre.}ITFuse
& 2.692 & 28.805 & 31.191 & 7.068 & 2.661 & 28.562 & 31.889 & 7.153 & 1.797 & 19.347 & 21.185 & 4.559 & 2.380 & 25.674 & 28.808 & 6.037 \\
\textit{pre.}Text-IF
& 4.175 & 43.569 & 40.228 & 12.147 & 3.924 & 41.107 & 41.138 & 12.236 & 5.738 & 60.027 & 43.247 & 16.281 & 5.662 & 59.539 & 51.261 & 15.667 \\
\textbf{Ours}     
& \textbf{4.497} & \textbf{46.715} & \textbf{41.271} & \textbf{13.518} & \textbf{4.384} & \textbf{45.745} & 41.455 & \textbf{13.378} & \textbf{6.939} & \textbf{72.514} & \textbf{51.834} & \textbf{19.324} & \textbf{6.724} & \textbf{70.366} & \textbf{52.654} & \textbf{18.743}\\
\bottomrule
\end{tabular}
}
\caption{
Quantitative comparison of degraded image fusion on the EMS dataset using VGDCFusion and seven comparative methods across four combined degradation scenarios (visible degradation \& infrared degradation). $\textit{pre.}$ denotes preprocessing applied to both source images (same as in the qualitative comparison). For Text-IF, the official model is used to handle visible degradation, while IR degradation is addressed via preprocessing. $\uparrow$ indicates that higher scores denote better performance. The best-performing results for each metric are highlighted in bold.}
\label{tab:Quantitative_Degrade}
\end{table*}

\subsection{Image Fusion with Degrade Guidance}
Existing methods typically assume high-quality inputs. To comprehensively evaluate the advantage of our approach on degraded image fusion, we preprocess degraded images using enhancement techniques before feeding them into fusion networks. For methods like Text-IF that are designed for single-type degradation, we additionally remove degradation from the other modality for fairness. Specifically, we apply MIRNet \cite{Zamir2022MIRNet} for enhancing infrared contrast, Swin-IR \cite{liang2021swinir} for infrared denoising, and IAT \cite{Cui_2022_IAT} for adjusting the brightness of visible images, thereby simulating and fairly comparing fusion under various combined degradation scenarios.

\subsubsection{Qualitative Analysis:}
The qualitative comparison of our method and seven state-of-the-art approaches on degraded image fusion is presented in Fig.~\ref{Qualitative_Degrade}. Across four degradation scenarios, competing methods exhibit various limitations:(1) \textbf{Low-light \& Low-contrast:} Except for Text-IF, visible enhancement is mostly effective, however, some visible details are lost in the fused results (green box, first row). All methods except CSF show insufficient contrast, particularly in textual regions (red box, first row). (2) \textbf{Low-light \& Noise:} All methods but Text-IF suffer from severe color distortions. Text-IF, while avoiding color shifts, fails to enhance visible light sufficiently, resulting in missing edge details (green box, second row). (3) \textbf{Overexposure \& Low-contrast:} GANMcC, SDNet, CSF, RFN-Nest, and Text-IF exhibit excessive inheritance or unnatural transitions in the infrared background, yielding low-quality fused backgrounds (red box, third row). DATFuse and ITFuse produce overly bright but low-contrast outputs under high-intensity conditions. All methods struggle to preserve salient targets (green box, third row). (4) \textbf{Overexposure \& Noise:} Similar degradation patterns reoccur, including unnatural fusion backgrounds and low-contrast results. Moreover, denoising leads to loss of infrared edge details (e.g., horizontal wires in red box, fourth row). In contrast, our method—through deep coupling of degradation modeling and fusion, along with prompt-guided degradation awareness—consistently generates clean, detail-rich, and visually compelling fusion results under various degradation conditions.

\begin{figure*}[!t]
\centering
\includegraphics[width=0.9\textwidth]{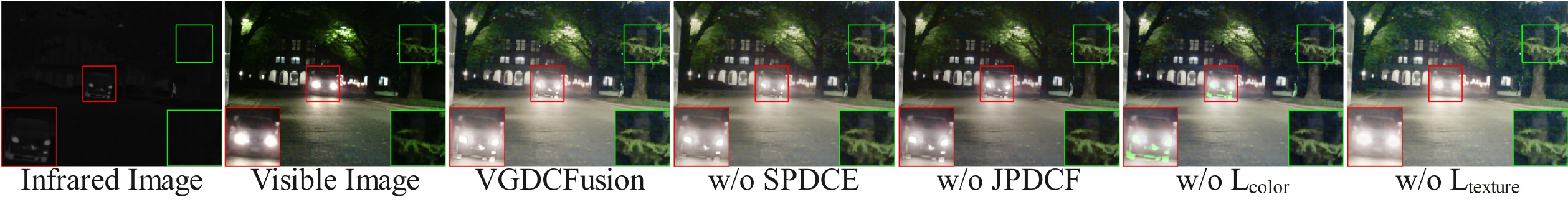}
\caption{
Qualitative analysis of VGDCFusion ablation studies. Key regions are highlighted with red and green boxes and magnified for clearer visual comparison.}
\label{Qualitative_Ablation}
\end{figure*}

\subsubsection{Quantitative Analysis:}
The quantitative comparison of our method against seven comparative approaches across four degraded fusion scenarios is shown in Tab.~\ref{tab:Quantitative_Degrade}. Except for a slightly lower performance in the $SD$ metric under the Low-light \& Noise condition (where DATFuse marginally outperforms), our method achieves superior scores across all remaining metrics and degradation scenarios. This further highlights the robustness and effectiveness of our network in handling degraded image fusion tasks.

\begin{table}[!t]
\centering
{\small
\begin{tabular}{ccccccc}
\toprule
 &\textbf{AG$\uparrow$} & \textbf{EI$\uparrow$} & \textbf{SD$\uparrow$} & \textbf{SF$\uparrow$}\\
\midrule
 w/o SPDCE & 4.140 & 43.025 & 38.586 & 12.591 \\
 w/o JPDCF & 4.304 & 44.800 & 39.338 & 12.932 \\
\midrule
 w/o $L_{color}$ & 4.399 & 45.711 & \textbf{42.199} & 13.511 \\
 w/o $L_{texture}$ & 4.107 & 43.011 & 41.679 & 12.016 \\
 
\midrule
\textbf{$\star$VGDCFusion}
             & \textbf{4.497}
             & \textbf{46.715}
             & 41.271
             & \textbf{13.518}\\
\bottomrule
\end{tabular}
}
\caption{
Quantitative analysis of VGDCFusion ablation studies. The best-performing results for each metric are highlighted in bold, and $\uparrow$ indicates that higher values correspond to better performance.
}
\label{tab:Quantitative_Ablation}
\end{table}

\subsection{Ablation Study}
\subsubsection{Qualitative Analysis:}
Fig.~\ref{Qualitative_Ablation} presents the qualitative results of the ablation study on the network architecture and loss functions of VGDCFusion. (1) Without SPDCE: The network's ability to extract modality-specific features is compromised, leading to weakened infrared intensity representation, as seen in the tire area (red box). (2) Without JPDCF: The lack of JPDCF impairs the degradation filtering, resulting in visible structural information loss such as the blurred tree trunk (green box). (3) Without $L_{color}$: The fused image suffers from noticeable color distortion, particularly evident in the tire region (red box). (4) Without $L_{texture}$: Although the overall brightness of the fusion result improves, substantial edge information is lost (red box).

\subsubsection{Quantitative Analysis:}
Tab.~\ref{tab:Quantitative_Ablation} presents quantitative ablation results of VGDCFusion's architecture and loss components. Removing SPDCE or JPDCF leads to consistent drops across all four metrics, indicating their critical role in performance. Excluding any loss term slightly improves $SD$ but degrades the other metrics, demonstrating the necessity of each loss function for optimal fusion quality.

\subsection{Application to Downstream Task}
\begin{figure}[!t]
\centering
\includegraphics[width=0.95\columnwidth]{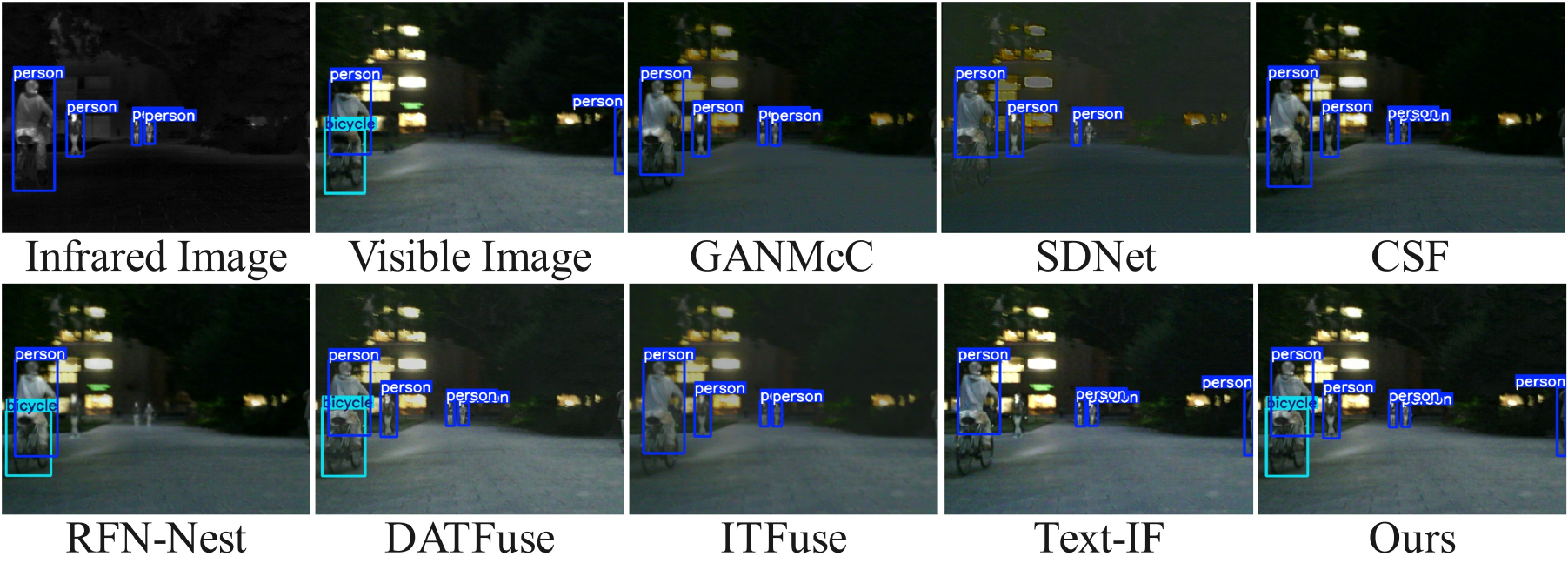}
\caption{Qualitative comparison of VGDCFusion and seven comparative image fusion methods on the MSRS dataset for the object detection task.}
\label{Qualitative_Detection}
\end{figure}

To assess VGDCFusion's effectiveness in downstream tasks, we fused 80 labeled MSRS image pairs and performed object detection using YOLOv5 on the results. \textbf{Qualitative Analysis:}
As shown in Fig.~\ref{Qualitative_Detection}, VGDCFusion outperforms single-modality and other fusion methods by accurately detecting diverse targets, including a bicycle, pedestrians at varying distances, and a partially occluded object.
\textbf{Quantitative Analysis:}
As shown in Tab.~\ref{tab:Quantitative_Detection}, VGDCFusion consistently surpasses single-modality and other fusion methods in mean average precision (mAP) across various intersection over union (IoU) thresholds, highlighting its effectiveness in downstream detection.

\begin{table}[!t]
\centering
{\small
\begin{tabular}{cccc}
\hline
\textbf{} & \textbf{mAP@0.65} &  \textbf{mAP@0.85} & \textbf{mAP@[0.5,0.95]} \\
\hline
IR       & 0.758 &  0.420 & 0.588 \\
VI        & 0.743 &  0.355 & 0.553 \\
GANMcC  & 0.842 &  0.534 & 0.656 \\
SDNet       & 0.795 &  0.487 & 0.641 \\
CSF        & 0.828 &  0.508 & 0.641 \\
RFN-Nest   & 0.775 &  0.476 & 0.593 \\
DATFuse    & 0.843 & 0.497 & 0.645 \\
ITFuse   & 0.839 & 0.493 & 0.640 \\
Text-IF & 0.823 & 0.491 & 0.644 \\
Ours      & \textbf{0.847} &  \textbf{0.569} & \textbf{0.665} \\
\hline
\end{tabular}
}
\caption{Quantitative comparison of VGDCFusion and seven comparative image fusion methods on the MSRS dataset for the object detection task. The best-performing results for each metric are highlighted in bold.}
\label{tab:Quantitative_Detection}
\end{table}

\subsection{Conclusion}
This paper proposes a novel degradation-aware image fusion network, termed \textbf{V}LM-\textbf{G}uided \textbf{D}gradation-\textbf{C}oupled Fusion (VGDCFusion), which addresses key limitations in existing degraded image fusion strategies—namely, the cumbersome switching between preprocessing methods and the performance degradation caused by the decoupling of degradation removal and image fusion. As the two core components of the network, the Specific-Prompt Degradation-Coupled Extractor (SPDCE) enables intra-modal degradation awareness and couples degradation suppression with feature extraction within each modality. Meanwhile, the Joint-Prompt Degradation-Coupled Fusion (JPDCF) facilitates cross-modal degradation perception and further establishes a coupling between degradation feature filtering and complementary feature aggregation across modalities. Extensive comparative experiments demonstrate that VGDCFusion achieves superior performance under various degraded scenarios compared to state-of-the-art methods.

\bibliography{aaai2026}
\end{document}